\documentclass{article} % For LaTeX2e
\usepackage{iclr2026_conference,times}

\usepackage[utf8]{inputenc}
\usepackage[T1]{fontenc}
\usepackage{hyperref}
\usepackage{url}
\usepackage{graphicx}

\iclrfinalcopy % This is a non-anonymous update to an existing arXiv preprint, not a blind conference submission.

\title{Patient-Centered Data Science: An Integrative Framework for \\ Evaluating and Predicting Clinical Outcomes in the Digital Health Era}

\author{Mohsen Amoei \\
McGill University Health Center Research Institute \\
Montreal, Quebec, Canada \\
\texttt{mohsen.amoei@mail.mcgill.ca} \\
\And
Dan Poenaru \\
McGill University Health Center Research Institute \\
Montreal, Quebec, Canada \\
\texttt{dan.poenaru@mcgill.ca}
}

\begin{document}

\maketitle

\begin{abstract}
This study proposes a novel, integrative framework for patient-centered data science in the digital health era. We developed a multidimensional model that combines traditional clinical data with patient-reported outcomes, social determinants of health, and multi-omic data to create comprehensive digital patient representations. Our framework employs a multi-agent artificial intelligence approach, utilizing various machine learning techniques including large language models, to analyze complex, longitudinal datasets. The model aims to optimize multiple patient outcomes simultaneously while addressing biases and ensuring generalizability. We demonstrate how this framework can be implemented to create a learning healthcare system that continuously refines strategies for optimal patient care. This approach has the potential to significantly improve the translation of digital health innovations into real-world clinical benefits, addressing current limitations in AI-driven healthcare models.
\end{abstract}

\section{Introduction}

The recent explosion in digital health has engendered claims of improved health outcomes through data science, wearable sensors, and artificial intelligence (AI) -- particularly the new large language models (LLMs). Despite numerous published models showing equivalent or even improved accuracy \emph{in vitro} compared to clinicians, very few AI models have been successfully integrated into clinical practice \citep{han2024randomised,wilkinson2020time}, with the first model successfully resulting in actual improved patient outcomes being a standard machine learning (ML) model \citep{lin2024aienabled}. This commentary explores the reasons behind the limited real-world progress in digital health and proposes a framework to address contemporary gaps and biases in digital health.

\section{Limitations in Legacy Evaluative and Prognostic Studies}

For centuries, clinical investigators have attempted to generate data which are accurate and faithful to their patients, for evaluating clinical interventions and predicting outcomes. Legacy clinical datasets have perennially been limited in both size and breadth, constrained by manual human collection abilities. These datasets were typically limited to demographic variables and clinical/physiological data such as signs and symptoms, laboratory results, and imaging data.

In recent decades, several other healthcare domains have become increasingly sampled, including genomic data, patient-centered outcomes, and a multitude of social determinants of health (SDoH). Integrating these new domains within ``traditional'' medical data introduced the challenges of platform incompatibilities and large dataset analysis by conventional statistical methods. Moreover, even the rare integrative biopsychosocial studies seldom addressed the various intersectional identities of the participants \citep{bauer2021intersectionality}. Even when broad healthcare domains were successfully combined with classic clinical datasets, personalized (as opposed to population-based) outcomes often remained elusive due to the loss of analytic power.

\section{The Data Science / Digital Health / AI Era}

Data science as a healthcare research space emerged in the 2000s, promising that ``big data'' would answer healthcare questions better and faster. The healthcare data flood began after the turn of the century with the widespread use of electronic health records (EHRs) and computerized monitoring devices, first introduced in the 1970s, followed by the rapid proliferation of wearable monitoring devices in the 2010s. Despite much industry-generated enthusiasm that large amounts of personal health data would improve overall diagnostic and therapeutic abilities, there are several key reasons why this prophecy remains largely unfulfilled.

\begin{enumerate}
\item The main reason is that our healthcare data, no matter how large or ``rich'', remains woefully incomplete. It is our assertion that, in fact, in the recent digital paradigm the extent of data incompleteness has increased, rather than decreased. Current healthcare datasets are continuously growing in size (latest global estimates placing them around 2 zettabytes, i.e. $2 \times 10^{21}$ bytes \citep{ellis2024kythera}) through digital data streams, yet their clinical spectrum rarely expands. Once data scientists became the collectors and curators of healthcare data, basic tenets such as ``bigger is better'' and ``any data is good data'' prevailed, leading to fierce competitions for private data and models gauged by computer science accuracy metrics rather than by patient outcomes \citep{arora2023value}.

\item Legacy datasets were primarily clinical and physiological, while current datasets are mostly physiologic and genomic. But precise, personalized care requires more than adding `omics and wearable data to basic clinical data -- it requires a solid sampling of broad other data domains/dimensions \citep{wilkinson2020time}. Even within the clinical data collected, the data science trend has been to categorize patients into simple binary categories (disease/feature present/absent), rather than within the continuum of risk so evident to clinicians \citep{wilkinson2020time}. In the absence of broad multidimensional data, the industry solution appears to be multimodal data -- combining text, images, audio, and video -- which unfortunately still only capture a very narrow human spectrum. Moreover, most current healthcare AI models are wholly proprietary, based on undisclosed datasets and factor weights, and hence prone to multiple, unknown, biases \citep{abramoff2023considerations}. Rather than somehow lessening bias through objectivity, there is ample evidence now that AI perpetuates -- and even augments -- existing biases in the data, effectively adding incremental bias through each phase of the total product lifecycle \citep{abramoff2023considerations}.

\item The data science promise was based not only on big data, but also on fast, complex analytical AI methods including machine learning, deep learning, and generative AI. The caveat of this promise is that analyzing big incomplete data using AI may create accurate \emph{in silico} models, but does not result in actual improved diagnostic and therapeutic abilities. Another key flaw of healthcare analytics is their inability to make causal inferences on the individual patient level (the very aim of precision/personalized medicine), being only able to classify patients and predict population-level outcomes \citep{wilkinson2020time}. Ultimately therefore, most industry-driven competition based on model predictive accuracy metrics has little relevance to clinical settings, where personalized measures of uncertainty are much more useful to clinicians than population-level accuracy \citep{banerji2023clinical}.

\item The industry's promise that algorithmic models developed on large proprietary patient datasets can be applied to one's own patient population is built on the ``myth of generalizability'' and the false assumption that external geographical validation is sufficient for generalizability -- when in fact multiple other threats to generalizability co-exist (changes in practice patterns, patient demographics, and SDoH over time) \citep{futoma2020myth}.
\end{enumerate}

When, despite the limitations listed above, healthcare researchers trust AI tools as ``research colleagues'', they become vulnerable to false illusions of explanatory depth and exploratory breadth \citep{messeri2024artificial}. These illusions, compounded by the model-induced belief revision seen in clinicians working with AI models \citep{kwong2024model}, threaten the scientific integrity of AI research, with patients ultimately being the ones to suffer.

\section{Suggested Solutions}

There are no clear, easy solutions to the current challenges in digital health. Any hope for a solution must however pause the race for more technology-reported outcomes (TechOs) \citep{mayo2017montreal}, and focus on the meticulous acquisition of truly multidimensional, real-life, longitudinal clinical datasets. These datasets must be patient-centered (rather than disease- or technology-centered), focus on health-related quality of life (HRQoL), and thus be anchored on patient-reported outcomes (PROs) and experiences (PREs). Outcomes reported by patients should not be clinician-defined (as in most generic and disease-specific PRO measures), but rather patients with lived experience of their illness should identify the outcomes that matter to them -- as in the rare so-called individualized PROMs, such as the Patient-Generated Index \citep{mayo2017support}.

The data must be multidimensional and multi-omic \citep{soenksen2022integrated}, including the various subcellular `omes (such as the genome, transcriptome, proteome, metabolome, etc.) and the full phenome (clinical data), but also the exposome, the sociome, and even measures of allostatic stress \citep{logan2018importance}. Longitudinal data is essential, considering patients' ongoing allostatic stress and its incremental impact on both wellness and illness trajectories. Within the exposome and sociome, SDoH and other stressors must be analyzed not only longitudinally but also intersectionally, capturing how social and political identities (such as gender, race, class, sexuality, disability, etc.) intersect and create unique experiences of privilege and oppression \citep{bauer2019advancing}. The recent concept of the social exposome \citep{gudimindermann2023integrating}, integrating social determinants and stressors within the broad range of environmental health exposures on an individual, aptly combines the principles of multidimensionality, reciprocity, and continuity along the life course. Beyond accurately describing various populations, the ultimate goal of integrating such vast, diverse patient data is the generation of highly granular, faithful digital representations of individual patients -- the ``digital twin'' concept invented decades ago -- and its use in the transformation of healthcare processes towards what Eric Topol dubbed ``deep medicine'' \citep{topol2019deep}.

Analyzing such complex, multi-omic datasets will require thoughtfully designed AI models beyond traditional statistics and simple ML models. LLMs appear well-suited for extracting unstructured textual EHR, survey, and qualitative data. Deep neural networks are able to explore complex multi-dimensional, multimodal, longitudinal data to discover latent connections, clustering, and intersectionalities.

\begin{figure}[t]
\begin{center}
\includegraphics[width=\linewidth]{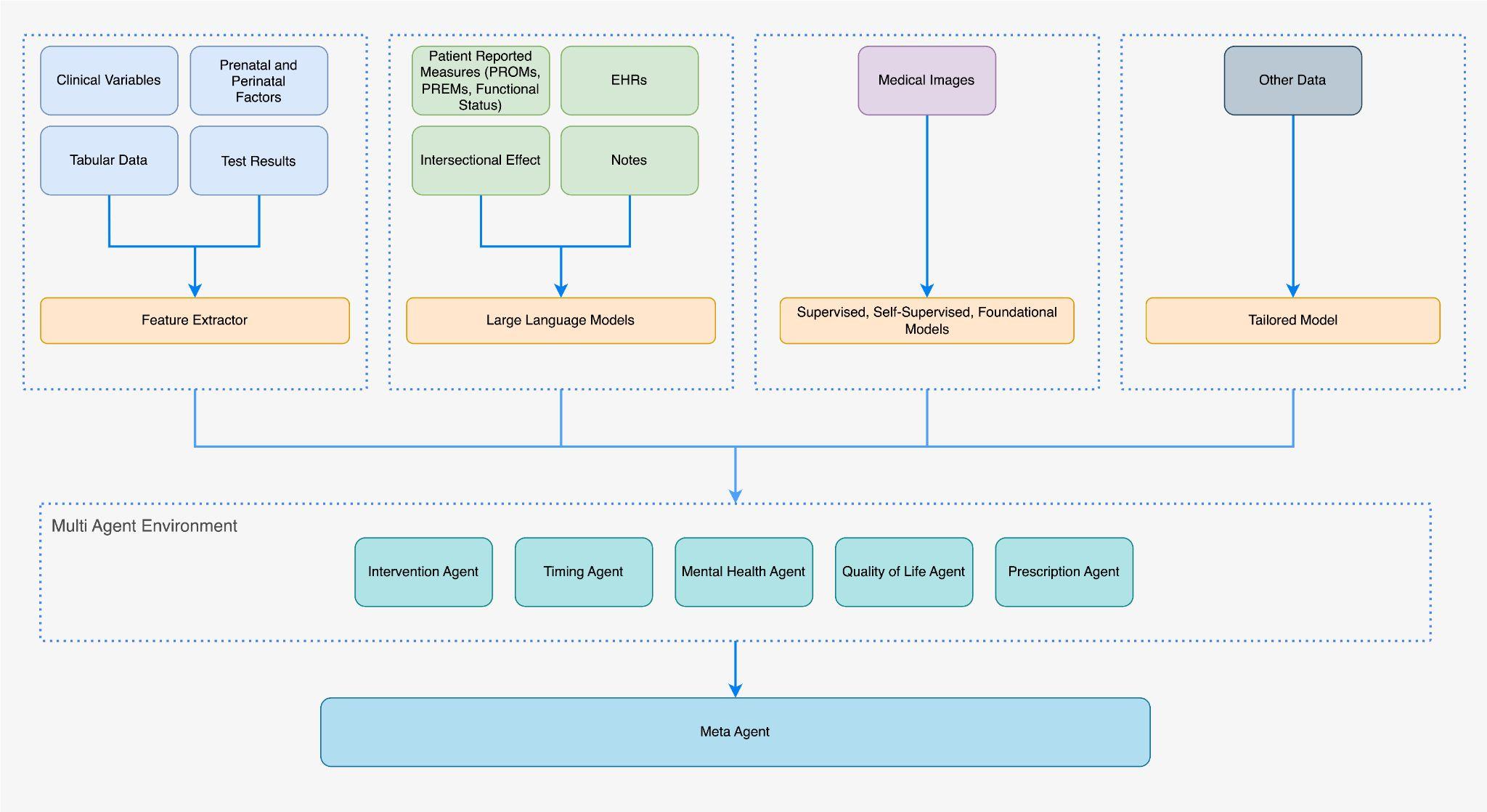}
\end{center}
\caption{Multi-dimensional, multimodal, multi-agent model.}
\label{fig:multiagent}
\end{figure}

To be truly patient-centric, healthcare AI models must reflect the multiple outcomes -- sometimes conflicting -- being sought within each population or sample. ``Improved healthcare'' as a model policy needs to be replaced by more precise outcomes, such as ``decreased mortality'', ``decreased short-term morbidity'', ``improved long-term clinical outcomes'', ``improved quality of life'', as well as ``decreased (or at least efficient) resource utilization'', to name a few. Within the overall AI model, each individual outcome is assigned to one AI agent, whose task is to optimize for that outcome. Once these individual agents have generated their optimized recommendations, a meta-agent aggregates and analyzes these suggestions to determine the best overall course of action for the patient (Figure~\ref{fig:multiagent}). The meta-agent, working under the supervision of a ``human-in-the-loop'' clinician, considers the interplay between various recommendations, ensuring a holistic approach to patient care that balances different aspects and priorities of the treatment. The meta-agent therefore not only synthesizes the insights from multiple specialized agents, but also adapts dynamically to the patient's evolving health status, embodying a learning healthcare system that continuously refines its strategies for optimal patient outcomes.

The individual agents adopt whichever AI approach best suits the task and the dataset characteristics. This may include both supervised and unsupervised learning, reinforcement learning (RL), and transformer-based approaches based on large-language models (LLMs). While traditional machine learning and deep learning approaches are best suited for tabular, quantitative data derived either from physiologic, genomic, and other tabular data, LLMs are particularly effective in extracting, summarizing, and analyzing unstructured and qualitative textual data. This unique ability to understand and interpret patient narratives and experiences through LLMs ensures that the model recommendations are not only data-driven, but also deeply aligned with the patients' subjective perspectives and needs. In light of the current concerns regarding the accuracy and reliability of LLMs, all such models must be fully anchored to the full medical corpus, and augmented and fine-tuned to the local datasets. While current approaches are still ill-defined in this emerging space, domain-specific retrieval-augmented generation \citep{zakka2024almanac}, transfer or reinforcement learning, and locally generated knowledge graphs \citep{gilbert2024augmented} are all promising venues.

Another key principle in the use of AI for healthcare data analysis, regardless of model used, is the intentional, assiduous search, exposure, and hopefully remediation of the countless biases threatening every step in each analytic pipeline. This starts with the very cautious use of pre-trained large language and foundational models, particularly proprietary ones based on undisclosed datasets and with unexamined biases. Models generated locally, on the very populations on which they will be applied, are ideal. Further steps towards creating bias-free models include continuous analyses of model outputs for potentially vulnerable populations and health determinants, and continuous model training and adjustments for temporal data shifts, data drifts, and therapeutic effects. This type of continuous internal validation is superior to the current external validation standards \citep{youssef2023external}, and can constitute the basis for a true learning healthcare system, a continuous cycle of data collection, analysis, intervention, and model improvement around the patient and their family.

Finally, the entire end-to-end AI healthcare pipeline must be evidence-based in its creation and continuous improvement. Ongoing, robust, comparative effectiveness studies of potential interventions must demonstrate their superiority or at least equivalence, ideally paired with decreased resource utilization. In keeping with the framework's fierce patient-centeredness, improved clinical outcomes, rather than technological model metrics, are to be sought and expected at all times. In this framework the healthcare providers, rather than industry interests and agents, will, fully partnered with their patients, drive the healthcare AI process.

\section{Proposed Framework}

\begin{figure}[t]
\begin{center}
\includegraphics[width=\linewidth]{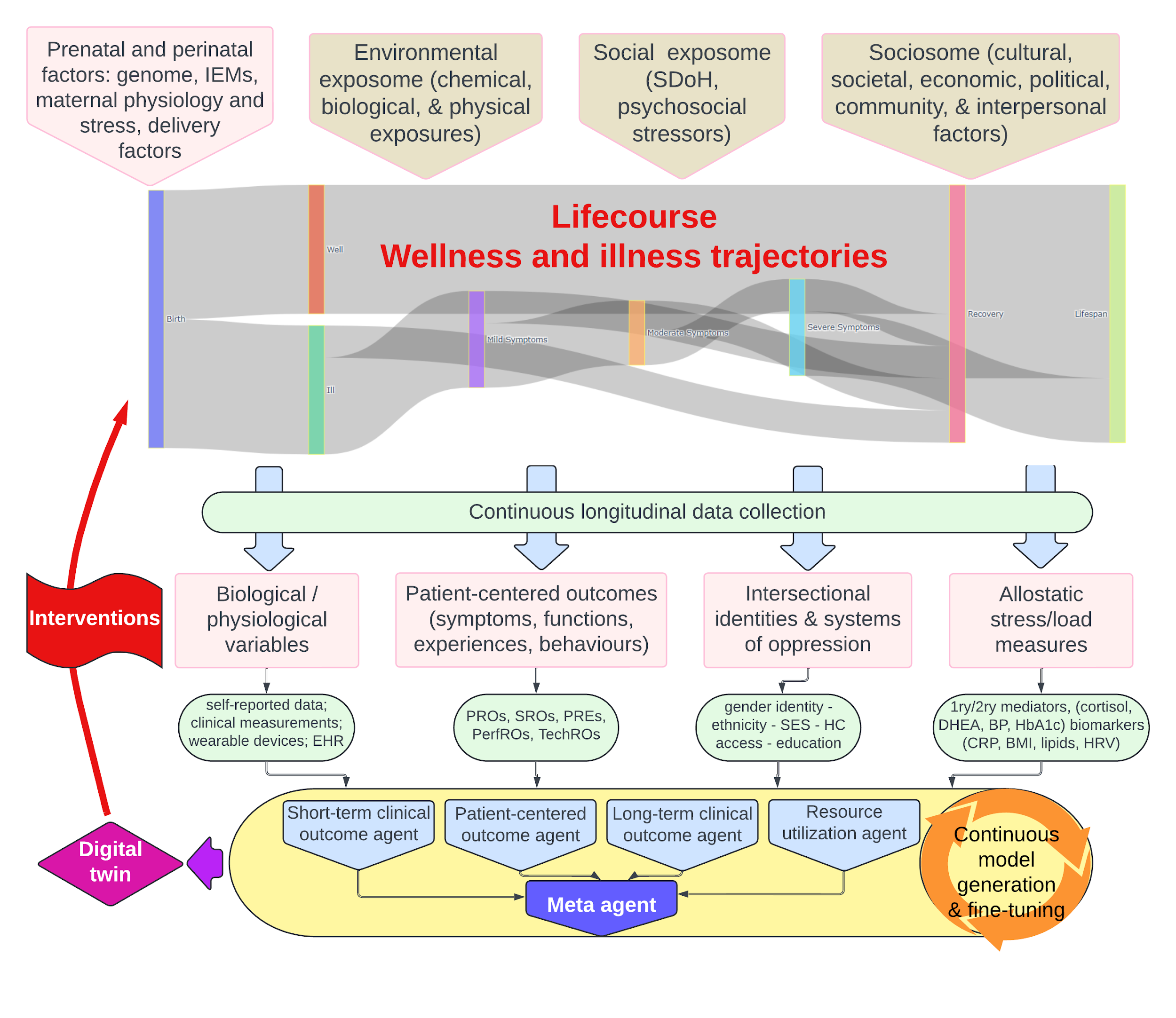}
\end{center}
\caption{Complete proposed framework for a patient-centered data science platform. SDoH: social determinants of health; IEMs: inborn errors of metabolism; EHR: electronic health record; PROs: patient-reported outcomes; SROs: self-report outcomes; PREs: patient-reported experiences; PerfROs: performance-reported outcomes; TechROs: technology-reported outcomes; SES: socio-economic status; HC: healthcare; BP: blood pressure; CRP: C-reactive protein; BMI: body-mass index; HRV: heart rate variability; DHEA: dehydroepiandrosterone.}
\label{fig:framework}
\end{figure}

Existing frameworks such as the Wilson-Cleary Model \citep{wilson1995linking}, the International Classification of Functioning model \citep{who2024icf}, the Valderas-Alonso model \citep{valderas2008patient}, the intersectionality wheel \citep{who2020incorporating}, and the social exposome framework \citep{gudimindermann2023integrating} provide as many useful foundations. The proposed framework below (Figure~\ref{fig:framework}) integrates these existing models with the digital health ecosystem, while maintaining patient-centeredness, multi-dimensionality, a lifespan longitudinal perspective, and continuous learning and improvement.

\section{Conclusion}

The integration of AI and data science into healthcare has the potential to revolutionize patient outcomes. However, to fully realize this potential, a paradigm shift towards truly patient-centered, multidimensional data collection and analysis is essential. Current practices that prioritize data quantity over quality and rely on proprietary models have shown limited real-world success. By focusing on longitudinal, intersectional datasets that capture the full spectrum of patient experiences and outcomes, and by employing collaborative AI techniques that can handle such complex data, we can move towards a more personalized and effective healthcare system.

Moreover, the development and implementation of AI models must be transparent, continuously locally validated, and deeply integrated into clinical workflows with a clear emphasis on improving patient outcomes rather than merely enhancing technological metrics. The proposed integrative framework, which combines the strengths of existing models with a focus on patient-centered data, offers a pathway to achieve these goals. By prioritizing patient-reported outcomes, leveraging collaborative AI techniques, and ensuring rigorous validation, we can create a healthcare environment that not only advances technology but also genuinely improves patient care and quality of life.

Ultimately, the future of digital health depends on our ability to balance innovation with patient-centeredness, ensuring that the benefits of AI and data science are fully realized in the clinical setting. This requires a commitment to ethical practices, continuous learning, and a relentless focus on the needs and experiences of patients. Through such efforts, we can transform the promise of digital health into a reality that delivers on its potential to enhance health and well-being for all.

\bibliography{references}
\bibliographystyle{iclr2026_conference}

\appendix
\section{Appendix: Existing Frameworks}

\subsection{Wilson-Cleary Model}
\begin{figure}[h]
\begin{center}
\includegraphics[width=0.85\linewidth]{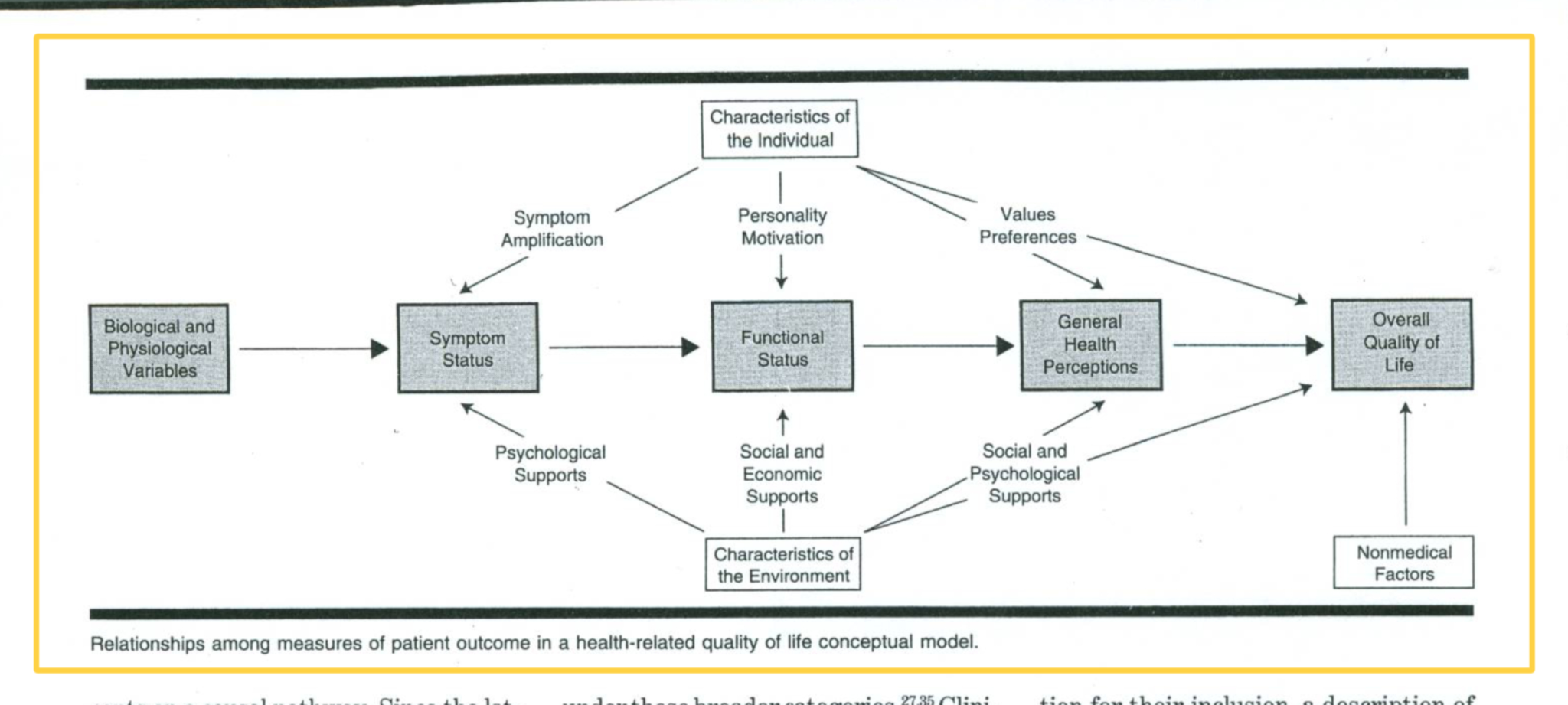}
\end{center}
\caption{Wilson-Cleary Model \citep{wilson1995linking}.}
\end{figure}

\subsection{Valderas-Alonso Model}
\begin{figure}[h]
\begin{center}
\includegraphics[width=0.85\linewidth]{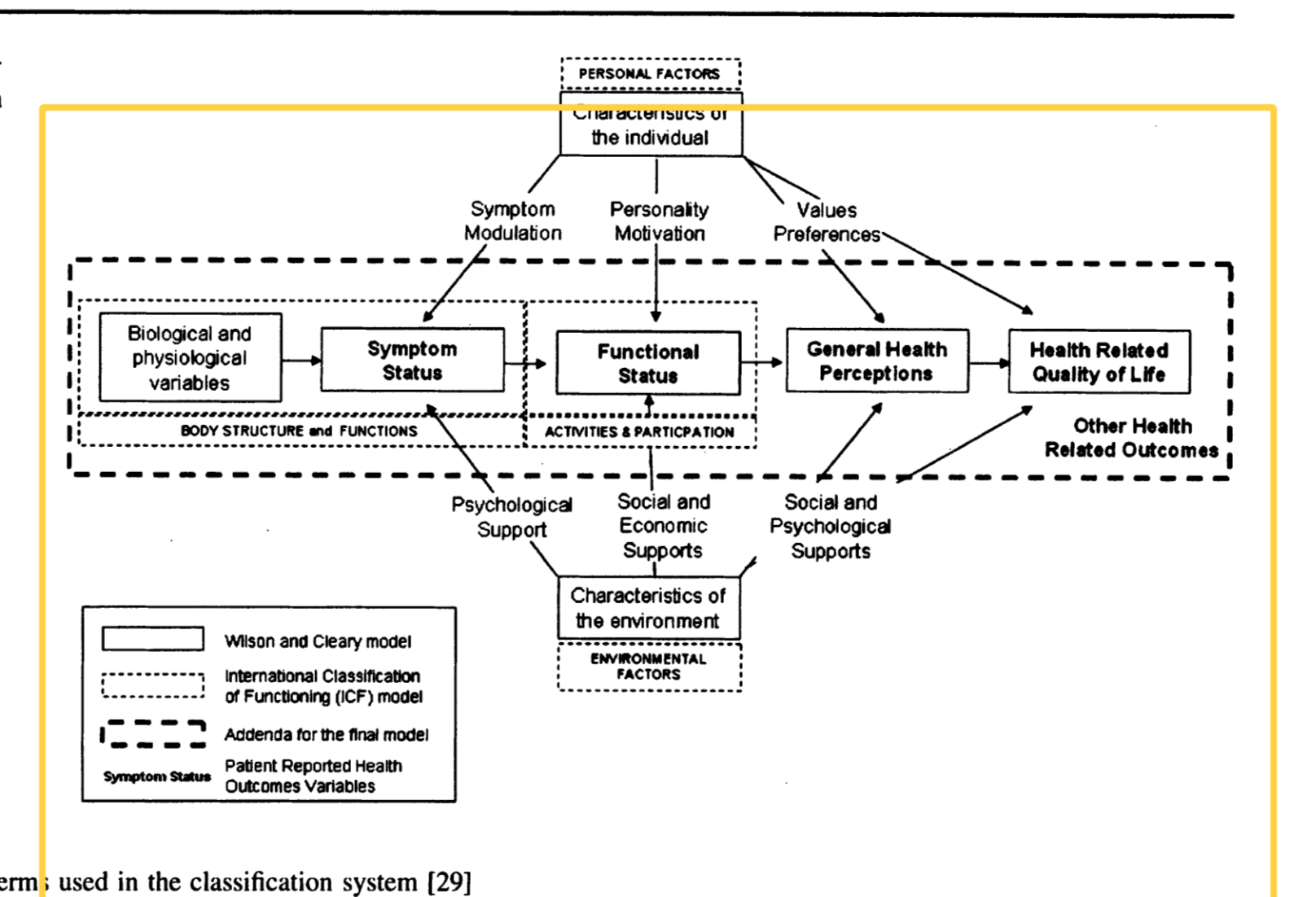}
\end{center}
\caption{Valderas-Alonso model \citep{valderas2008patient}.}
\end{figure}

\subsection{Intersectionality Wheel}
\begin{figure}[h]
\begin{center}
\includegraphics[width=0.85\linewidth]{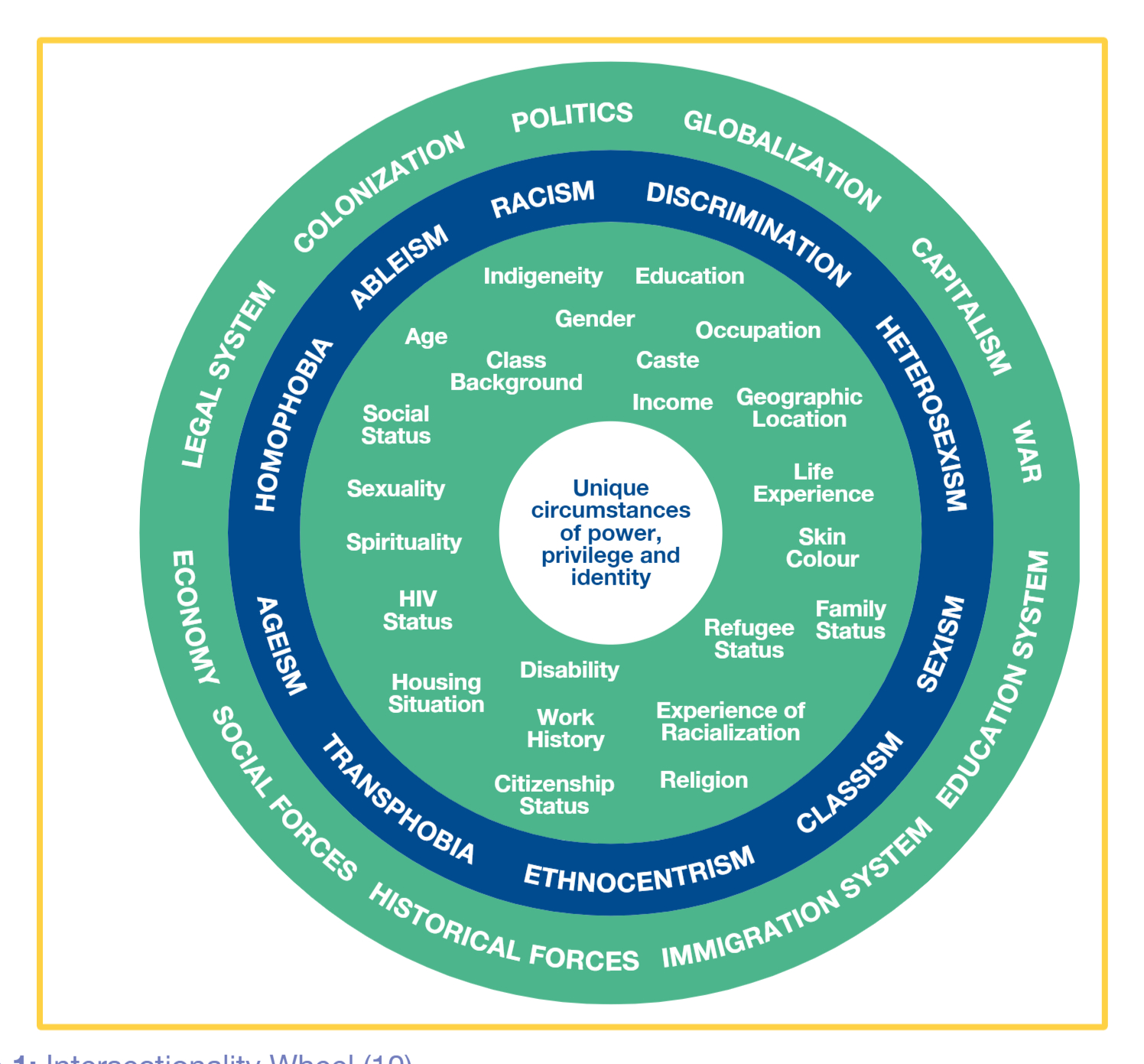}
\end{center}
\caption{Intersectionality wheel \citep{who2020incorporating}.}
\end{figure}

\subsection{Social Exposome}
\begin{figure}[h]
\begin{center}
\includegraphics[width=0.85\linewidth]{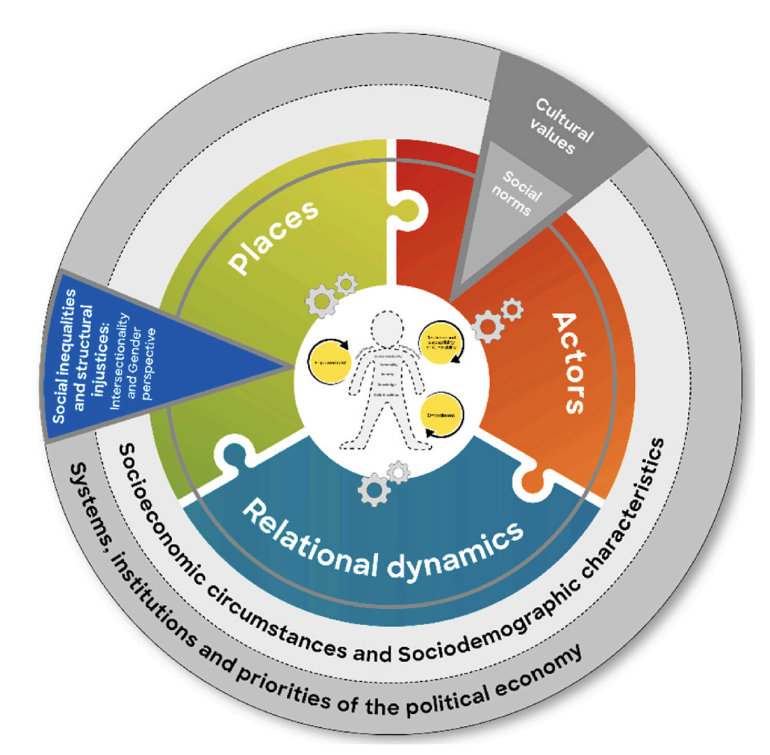}
\end{center}
\caption{Social Exposome \citep{gudimindermann2023integrating}.}
\end{figure}

\end{document}